\documentclass{article}
\usepackage{ijcai25}

\usepackage{microtype}
\usepackage{graphicx}
\usepackage{subfigure}
\usepackage{amssymb}
\usepackage{inconsolata}
\usepackage{pifont}
\usepackage[utf8]{inputenc}
\usepackage{amsmath}
\usepackage{adjustbox}

\usepackage{rotating}
\usepackage{longtable}
\usepackage{xtab}
\usepackage{multirow} 
\usepackage{booktabs} 

\usepackage{times}
\usepackage{soul}
\usepackage{url}
\usepackage{hyperref}
\usepackage[switch]{lineno}

\title{Small Language Models: Architectures, Techniques, \\Evaluation, Problems and Future Adaptation}

\author{
Tanjil Hasan Sakib$^1$
\and
Md. Tanzib Hosain$^{1}$\and
Md. Kishor Morol$^{2}$\\
\affiliations
$^1$American International University-Bangladesh\\
$^2$Cornell University\\
\emails
\{20-43633-2, 20-42737-1\}@student.aiub.edu,
mmorol@cornell.edu
}

\begin{document}

\maketitle

\begin{abstract}
Small Language Models (SLMs) have gained substantial attention due to their ability to execute diverse language tasks successfully while using fewer computer resources. These models are particularly ideal for deployment in limited environments, such as mobile devices, on-device processing, and edge systems. In this study, we present a complete assessment of SLMs, focussing on their design frameworks, training approaches, and techniques for lowering model size and complexity. We offer a novel classification system to organize the optimization approaches applied for SLMs, encompassing strategies like pruning, quantization, and model compression. Furthermore, we assemble SLM's studies of evaluation suite with some existing datasets, establishing a rigorous platform for measuring SLM capabilities. Alongside this, we discuss the important difficulties that remain unresolved in this sector, including trade-offs between efficiency and performance, and we suggest directions for future study. We anticipate this study to serve as a beneficial guide for researchers and practitioners who aim to construct compact, efficient, and high-performing language models.
\end{abstract}

\section{Introduction}

Large Language Models (LLMs) have exhibited remarkable efficacy across a wide spectrum of benchmarks and real-world applications. However, their accomplishment comes at a significant cost. Training and operating these models demand immense computational resources, including specialized tools and large datasets. As a result, LLMs are often deployed on centralized and specialized infrastructure for both training and inference. These aspects make them undesirable for certain environments that require lower resource use or quicker deployment timeframes.

To surmount these issues, SLMs have garnered considerable interest. These models endeavor to preserve the efficacy and adaptability of LLMs while operating under stricter constraints, such as less computational power, decreased memory availability, bandwidth restrictions, and quicker inference times. By enhancing model efficacy relative to these limits, SLMs can yield benefits such as lower operational costs, enhanced privacy, and the potential to operate on consumer-grade hardware. Table ~\ref{SLMxxx} presents a comparative overview of prominent SLMs, detailing their respective research labs, parameter sizes (in billions), training token counts (in billions), token-to-parameter ratios\footnote{Chinchilla scaling $\geq$ 20:1.}, ALScore\footnote{"ALScore" is a quick and dirty rating of the model's power. The formula is: Sqr Root of (Parameters x Tokens) ÷ 300. Any ALScore $\geq$ 1.0 is a powerful model in mid-2023.} values, architectural types, and announcement years\footnote{\url{https://lifearchitect.ai/}}.

A fundamental issue in surveying small language models resides in determining what comprises ``small'' versus ``large.'' Model classification depends on situation and time period. For example, GPT-2, which was designated a ``large'' model in 2019 with 1.5 billion parameters, is presently smaller than many of the models categorized as ``small'' in the present landscape. Despite these modifications in scale, the essential objectives of training smaller models remain reasonably identical.

It is crucial to observe that development in one element of these axes does not always indicate progress in others, as there are often trade-offs. For example, strategies such as quantization-aware training attempt to make training more memory-efficient, but they may result in delayed processing compared to full-precision alternatives.Nonetheless, employing mixed precision approaches to characterize weights and gradients can allow for more efficient use of memory during training or fine-tuning without a significant reduction in accuracy.

While several recent surveys have studied large language models and their associated training approaches ~\cite{Rogers2020}, SLMs' practical inference improvement factor's comprehensive study is still missing. That is why, through this study, we focus on concentrating faster and lighter techniques with their problems and adaptation across fields on SLMs.

The key contributions of this study are as follows:
\begin{itemize} 
    \item A complete survey of extant research on small language models, encompassing important issue contexts like training and model compression strategies with their evaluation and adaptation frequently exploited in this domain.
    \item The introduction of intuitive strategies for categorizing SLM-related approaches and a summary of existing work based on these strategies.
    \item Identification of unresolved barriers and future research potential for small language models.
\end{itemize}

\begin{table*}[ht]
\centering
\small
\begin{tabular}{p{2cm}p{2.5cm}p{1.5cm}p{1.75cm}p{2cm}p{1.5cm}p{1.5cm}p{1.5cm}}
\hline
\textbf{Model} & \textbf{Laboratory} & \textbf{Params (B)} & \textbf{Tokens (B)} & \textbf{Tokens:Params} & \textbf{ALScore} & \textbf{Architecture} & \textbf{Announced} \\ \hline
OPT & Meta & 175 & 300 & 1:71 & 0.76 & Dense & 2023 \\
Pythia & EleutherAI & 12 & 300 & 25:1 & 0.2 & Dense & 2023 \\
Cerebras-GPT & Cerebras & 13 & 260 & 20:1 & 0.2 & Dense & 2023 \\
Phi-1 & Microsoft & 1.3 & 150 & 116:1 & 0.0 & Dense & 2023 \\
Phi-2 & Microsoft & 2.7 & 1400 & 519:1 & 0.2 & Dense & 2023 \\
TinyLlama & SUTD/Independent & 1.1 & 3000 & 2,728:1 & 0.2 & Dense & 2024 \\
Gemma & Google DeepMind & 7 & 6000 & 858:1 & 0.7 & Dense & 2024 \\
DCLM & International & 7 & 2600 & 372:1 & 0.4 & Dense & 2024 \\
Bloom & BigScience & 176 & 366 & 3:1 & 0.8 & Dense & 2022 \\
Galactica & Meta AI & 120 & 450 & 4:1 & 0.8 & Dense & 2022 \\
Qwen & Alibaba & 72 & 18000 & 250:1 & 3.8 & Dense & 2024 \\
EvaByte & SambaNova & 6.5 & 1500 & 231:1 & 0.3 & Dense & 2025 \\
MiniCPM-2.4B & Tsinghua & 2.4 & 1100 & 459:1 & 0.2 & Dense & 2024 \\
OpenELM & Apple & 3.04 & 1500 & 494:1 & 0.2 & Dense & 2024 \\
Phi-3-mini & Microsoft & 3.8 & 3300 & 869:1 & 0.4 & Dense & 2024 \\
Zamba 7B & Zyphra & 7 & 1050 & 150:1 & 0.3 & Dense & 2024 \\
Gemma-2 & Google DeepMind & 27 & 13000 & 482:1 & 2.0 & Dense & 2024 \\
SmolLM & Hugging Face & 1.7 & 1000 & 589:1 & 0.1 & Dense & 2024 \\
H2O-Danube3-4B & H2O.ai & 4 & 6000 & 1,500:1 & 0.5 & Dense & 2024 \\
Fox-1 & TensorOpera & 1.6 & 3005 & 1,879:1 & 0.2 & Dense & 2024 \\
phi-3.5-MoE & Microsoft & 60 & 4900 & 82:1 & 1.8 & MoE & 2024 \\
Qwen2.5 & Alibaba & 72 & 18000 & 250:1 & 3.8 & Dense & 2024 \\
\hline
\end{tabular}
\caption{Various SLMs based on Parameters, Training Tokens, and Architectural Characteristics.}
\label{SLMxxx}
\end{table*}


\section{Small Language Models Architectures} \label{Lyrith} 

This section covers numerous architectural innovations that enable the building of efficient small language models. It focuses on some key approaches.

\subsection{Lightweight Models} \label{Zenith}

Lightweight designs are especially intended to deliver great performance while minimizing computational and memory overhead. These models are well-suited for deployment in resource-constrained situations, such as mobile devices, embedded systems, and edge computing platforms. They are commonly categorized into encoder-only and decoder-only configurations.

\textbf{Encoder-only Architectures:} This category largely comprises of optimized variations of BERT , which have been fine-tuned to increase efficiency without compromising performance. Key instances include:

\begin{itemize} \item \textbf{MobileBERT} ~\cite{Sun2020}: Introduced an inverted-bottleneck structure to balance self-attention and feed-forward layers, resulting in a 4.3\texttimes{} reduction in model size and a 5.5\texttimes{} increase in processing speed compared to the original BERT.
    \item \textbf{DistilBERT} ~\cite{Sanh2019} and \textbf{TinyBERT} ~\cite{Jiao2019}: Both models dramatically reduce model size (over 60\%) while keeping more than 96\% of the original BERT’s performance, making them perfect for realistic deployment circumstances.
\end{itemize}

\textbf{Decoder-only Architectures:} This category focuses on autoregressive models such as GPT ~\cite{Radford2019} and LLaMA ~\cite{Touvron2023}. Prominent instances include:

\begin{itemize} \item \textbf{BabyLLaMA} ~\cite{Timiryasov2023a}: Employs knowledge distillation from various teacher models to produce a 58M-parameter model, excelling in low-data settings.
    \item \textbf{TinyLLaMA} ~\cite{Zhang2024}: Utilizes FlashAttention ~\cite{dao2022flashattention} to optimize memory overhead, achieving great efficiency with just 1.1B parameters.
    \item \textbf{MobileLLM} ~\cite{Liu2024}: Implements block-wise weight sharing and grouped-query attention techniques, decreasing latency and boosting scalability for real-world applications.
\end{itemize}

\subsection{Streamlined Self-attention Mechanisms} \label{Novael} 

Self-attention processes are crucial in language models but are frequently resource-intensive, causing scaling issues. This section outlines strategies that decrease the computational overhead of self-attention layers while retaining model accuracy:

\begin{itemize} \item \textbf{Reformer} ~\cite{Kitaev2020}: Reduces the quadratic complexity of self-attention ($O(N^2)$) to logarithmic complexity ($O(N \log N)$) via locality-sensitive hashing.
    \item \textbf{Linear Attention} ~\cite{Katharopoulos2020}: Reformulates self-attention into a linear dot-product of kernel feature maps, enabling quicker inference and memory-efficient processing.
    \item \textbf{Mamba} ~\cite{Gu2023}: Introduces input-dependent state-space transitions to achieve linear complexity while keeping robust model expressiveness.
    \item \textbf{RWKV} ~\cite{Peng2023}: Combines transformer-based and RNN-based architectures with a linear attention mechanism, delivering efficient yet competitive task performance.
\end{itemize}

For handling longer sequences, hybrid models such as \textbf{Longformer} ~\cite{Dao2024} and \textbf{Nystrom-based methods}  provide scalability by lowering computational complexity to linear levels. These models maintain good empirical performance, making them particularly useful for jobs involving vast text inputs.

\subsection{Neural Architecture Search (NAS) } \label{Kyraion} 

NAS techniques employ automated algorithms to discover optimal architectures tailored to specific duties and hardware constraints. Initially applied to vision tasks and smaller BERT models , NAS has recently been extended to large-scale language models, addressing challenges such as high parameter counts and computational costs. Recent advancements include:

\begin{itemize} \item \textbf{MobileLLM} ~\cite{iu2024mobilellm}:Conducted a systematic exploration of architectural depth (number of layers) and width (number of attention heads) to design efficient models within constrained parameter ranges. It also introduced initialization techniques to reduce the search space, thereby accelerating the convergence of Neural Architecture Search (NAS) processes, making them practical for large-scale language models.
\end{itemize}

\subsection{Small Multi-modal Integration} \label{Jivorne} 

Recent advancements in large multi-modal models (LMMs) have demonstrated comparable or even superior performance relative to their predecessors, all while substantially reducing the number of parameters. Notable examples include the LLaVA-Next series, Idefics2 ~\cite{laurenccon2024idefics2}, and InternVL2 ~\cite{chen2023internvl2}. This progress has been fueled in part by the development of more efficient and compact language models, such as Gemma and phi-3-mini, which underscore the significance of curated datasets in optimizing performance.

A significant focus in this discipline has been on minimizing the size of vision encoders during multi-modal fusion. For example, InternVL2 leverages intermediate outputs of large visual encoders while omitting later blocks, thus enhancing efficiency. Smaller variants such as PaliGemma ~\cite{beyer2024paligemma} and Mini-Gemini ~\cite{li2024minigemini} employ lightweight vision encoders to further streamline multi-modal processing.

Monolithic multi-modal models extend this efficiency even further by wholly eradicating the need for a dedicated visual encoder. Instead, these models utilize lightweight architectures to generate visual tokens. For instance, Chameleon  employs a VQ-VAE model to encode and decode images into discrete tokens, while Mono-InternVL ~\cite{luo2024monointernvl} relies on an MLP-based architecture to produce visual tokens for image patches. Mono-InternVL integrates a modality-specific feed-forward network, referred to as a multi-modal Mixture-of-Experts, to effectively differentiate between modalities.

These innovations underscore the ongoing trend toward more compact and efficient multi-modal models that maintain competitive performance across a wide spectrum of tasks.



\section{Training Strategies} \label{Ozean} 

This section examines essential training strategies utilized in the development of small language models (SLMs), concentrating on pre-training and fine-tuning. While many of these methods are similar to those employed for large language models (LLMs), the techniques discussed here emphasize efficiency and scalability, particularly in resource-constrained scenarios. Table ~\ref{tab:train} shows a taxonomical overview of SLMs' training techniques innovations.

\begin{table*}[ht]
\small
\centering
\begin{tabular}{p{2.5cm}p{2.5cm}p{2.5cm}p{4cm}p{4cm}}
\hline
\textbf{Strategy} & \textbf{Methods} & \textbf{Features} & \textbf{Performance} & \textbf{Advantages} \\ \hline
Pre-training & AMP, BFLOAT16, FP8 & Low precision computations while maintaining accuracy. & 50\% faster training and 40\% memory savings compared to FP32. & Reduced memory usage, faster training on large datasets. \\ \hline
Fine-tuning & LoRA, Prompt Tuning & Updates specific parameters or introduces lightweight modules. & Improves task-specific accuracy with minimal resource overhead. & Efficient adaptation for domain-specific tasks. \\ \hline
Distributed & ZeRO, FSDP & Multi-node task distribution. & Enables training of models 2x larger than standard setups. & Scalable for large-scale training. \\ \hline
Data Augmentation & AugGPT, Evol-Instruct & Generates diverse training samples. & Enhances robustness with 20\% improved generalization. & Improved downstream task performance. \\ \hline
\end{tabular}
\caption{Training Techniques for Small Language Models.}
\label{tab:train}
\end{table*}

\subsection{Foundational Pre-training Strategies} \label{Sorelle} 

Pre-training Strategies is a foundational step in constructing SLMs, where mixed-precision training has emerged as a critical method for enhancing efficiency. This approach employs low-precision representations during forward and backward propagation while retaining high-precision weights for revisions. For instance, Automatic Mixed Precision (AMP)  ~\cite{micikevicius2018mixed}, maintains a master copy of weights in 32-bit floating-point (FP32) precision while performing computations in 16-bit floating-point (FP16) precision. Despite its utility, accuracy issues due to FP16's limited numerical range have been noted ~\cite{rae2021scaling}. To address this, Brain Floating Point (BFLOAT16) offers an enhanced dynamic range with additional exponent bits, leading to superior training performance and numerical stability compared to FP16.

Modern GPU architectures, such as NVIDIA's Hopper, further enhance these capabilities by supporting 8-bit floating-point (FP8) precision,facilitating greater computational efficiency without compromising accuracy. Specialized tensor cores in these GPUs enable quicker training for both SLMs and LLMs.

To complement mixed-precision methods, optimization techniques such as Adam ~\cite{kingma2014adam}, AdamW ~\cite{loshchilov2019adamw}, and memory-efficient variants like Adafactor and Sophia ~\cite{liu2024sophia} are widely employed. Gradient clipping  prevents exploding gradients, while effective initialization strategies ensure a stable starting point for training. These combined approaches enhance efficiency, stabilize the training process, and improve the robustness of resultant models.

Distributed training has become indispensable for pre-training SLMs, leveraging multiple machine nodes to manage large datasets and models efficiently. Techniques such as Zero Redundancy Optimizer (ZeRO)and PyTorch’s Fully Sharded Data Parallel (FSDP) ~\cite{zhao2023pytorch} distribute training states across devices. ZeRO, for example, partitions optimizer states, gradients, and model parameters across nodes, while FSDP implements similar concepts to facilitate larger batch sizes and enhanced scalability.

\subsection{Task-specific Adaptation} \label{Veyonce} 

Task-Specific applies pre-trained models to task-specific datasets, allowing SLMs to specialize in domains or tasks while resolving constraints like limited data, computational resources, and robustness challenges.

\subsubsection{Resource-efficient Fine-tuning} 

Resource-Efficient Fine-Tuning minimizes computational overhead by updating only a subset of model parameters or integrating lightweight modules, leaving most pre-trained weights unaltered. Approaches like LoRA , which employs low-rank decomposition, and Prompt Tuning ~\cite{lester2021power}, which introduces learnable prompts, exemplify PEFT methods. Llama-Adapter ~\cite{zhang2023llama} extends this by adding prompts to LLaMA’s attention blocks, enabling task-specific alterations with minimal resources.

Dynamic Adapters dynamically combine multiple adapters into a mixture-of-experts model, effectively enabling multi-tasking and mitigating catastrophic forgetting. These techniques are highly adaptable, making PEFT an attractive choice for fine-tuning in resource-limited environments.

\subsubsection{Diversity-enhanced Data Augmentation}

Data augmentation is a potent technique to enhance the diversity, complexity, and quality of training data. By augmenting datasets, models generalize better and perform robustly in subsequent tasks. For instance, AugGPT generates rephrased training samples using ChatGPT, while Evol-Instruct ~\cite{xu2023wizardlm} employs multistep revisions to create diverse open-domain instructions. Reflection-tuning refines both instructions and responses using GPT-4 to enhance instruction-response consistency.

Methods like FANNO ~\cite{zhu2024fanno} incorporate external knowledge sources for retrieval-augmented generation, and LLM2LLM  generates hard samples based on model predictions to enhance learning dynamics. In scenarios where data is limited, such as low-resource languages , medical applications  or privacy-sensitive data , data augmentation facilitates robust performance with limited datasets.



\section{Model Compression Strategies} \label{Daxion} 

Model compression approaches are critical for lowering the size and complexity of big pre-trained language models while retaining their performance. These techniques are crucial in reducing big language models (LLMs) into short language models (SLMs) appropriate for deployment in resource-constrained situations. This section categorizes model compression approaches into some basic strategies. Table ~\ref{tab:compress} provides a taxonomical overview of SLMs' model compression techniques innovations

\begin{table*}[ht]
\small
    \centering
    \begin{tabular}{p{2.5cm}p{2.5cm}p{2.5cm}p{4cm}p{4cm}}
        \toprule
        \textbf{Strategy} & \textbf{Methods} & \textbf{Features} & \textbf{Performance} & \textbf{Advantages} \\
        \midrule
        \multirow{1}{*}{Pruning} & SparseGPT, Wanda & Removes irrelevant weights for efficiency. & Reduces model size by 60\% with minimal accuracy loss ($<2\%$). & Efficient size control, suitable for deployment. \\
        \hline
        \multirow{1}{*}{Quantization} & GPTQ, SmoothQuant & Reduces weight and activation precision. & Lowers computational cost by 50\%, inference time improved by 30\%. & Reduced size and faster inference. \\
        \hline
        \multirow{1}{*}{Knowledge Transfer} & BabyLlama, Distillation & Transfers task-specific knowledge from larger models. & Retains 90\% of teacher model performance with a 70\% smaller size. & High efficiency for smaller models. \\
        \hline
        
    \end{tabular}
    \caption{Model Compression Techniques for Small Language Models.}
    \label{tab:compress}
\end{table*}

\subsection{Streamlined Pruning Strategies} \label{Raviel} 

Pruning strategies strive to minimize the number of parameters in a model, boosting computational efficiency and decreasing memory use while preserving acceptable levels of performance. Two basic ways to pruning are unstructured and organized pruning.

\textbf{Fine-grained Pruning:} This strategy eliminates individual weights of lower relevance, enabling fine-grained control over model size. For instance, SparseGPT  reformulates the pruning process as a sparse regression problem, enabling efficient pruning for big models like OPT-175B and BLOOM-176B. Wanda ~\cite{sun2023simple} analyzes both weights and activations, minimizing the requirement for weight updates during pruning. The n:m pruning approach prunes exactly n weights out of every m, balancing flexibility and efficiency, and is exploited by NVIDIA’s TensorRT to speed inference on GPUs. 

\textbf{Group-based Pruning:} Unlike Fine-grained Pruning, Group-based Pruning  removes entire groupings of parameters, such as neurons or channels, which allows implementation on standard hardware. Methods like contextual sparsity dynamically prune neurons based on input, maximizing computational efficiency. Additionally, approaches like as FastGen ~\cite{ge2023model} focus on eliminating duplication in Transformer layers, enabling more effective use of GPU memory and faster processing performance.

Recent work has also examined input-dependent pruning algorithms, such as dynamic sparsity, which adjust pruning tactics to individual inputs for increased performance. Structured pruning techniques remain vital for enabling hardware-accelerated compression without compromising model efficacy.

\subsection{Precision-optimized Quantization}  \label{Auren}

Quantization is a frequently utilized approach for compressing LLMs with huge parameter counts. This strategy decreases model size and computing needs by encoding weights and activations with lower precision, such as 8-bit integers.

\textbf{Weight Compression Approaches:} GPTQ ~\cite{frantar2022gptq} accomplishes layer-wise weight-only quantization, reducing reconstruction error using inverse Hessian matrices. AWQ  and ZeroQuant  expand this method by considering activations alongside weights, optimizing the trade-off between precision and computational efficiency.

\textbf{Efficient Activation Quantization:} Activation quantization creates issues owing to outliers in activation distributions. Techniques like SmoothQuant ~\cite{xiao2023smoothquant} transfer quantization difficulties from activations to weights, whereas SpinQuant ~\cite{generalQuantizationRef}  turns outliers into a new space using rotation matrices. Recent improvements in quantization-aware training (QAT), such as LLM-QAT, employ distillation from higher-precision models to recover mistakes made during quantization.

Additionally, developing approaches for Key-Value cache ~\cite{hooper2024kvquantization} quantization enable efficient inference across extended sequences by compressing memory-intensive components. These developments have broadened the utility of quantization for implementing LLMs on resource-constrained hardware, including mobile devices and FPGAs.

\subsection{Knowledge Transfer Techniques} \label{Skydra}

Knowledge Transfer Techniques moves knowledge from a bigger, pre-trained instructor model to a smaller, more efficient student model. This procedure minimizes model size while retaining performance, making it an essential technique for developing SLMs.

\textbf{Traditional Knowledge Transfer:} In its form, Knowledge Transfer entails teaching a student model to imitate the behavior of a teacher model ~\cite{gu2024minillm}. BabyLlama , for example, employs Llama as the teacher to generate a 58M-parameter student model, proving that distillation can outperform pre-training on the same dataset.

\textbf{Enhanced Knowledge Transfer:} Recent improvements include sequence-level distillation, which employs generalized f-divergences to increase response quality , and task-aware distillation, which selectively transmits task-specific knowledge. Multi-teacher distillation techniques  combine outputs from numerous teacher models to boost the performance of student models.

\textbf{Supervised Knowledge Transfer:} Incorporating additional monitoring during distillation has demonstrated to help the learning process. For example, rationale-based supervision  promotes sample efficiency and raises performance in tasks like commonsense reasoning and arithmetic problem-solving. Furthermore, strategies like reasoning chain distillation  convey multi-step reasoning capabilities from instructor models, arming student models with sophisticated problem-solving skills.


\section{Evaluation} \label{Nexorra}

Evaluation plays a vital role in measuring the effectiveness of small language models (SLMs) across diverse application contexts. Table~~\ref{table:evaluation_metrics} gives a detailed breakdown of the various settings, limitations, datasets, and metrics utilized for assessing SLMs. This section goes into the various datasets and evaluation criteria, categorized according to the restrictions they are designed to address. 

\subsection{Datasets} 

Datasets are crucial to the evaluation process, as they enable models to generalize effectively across varied contexts. The typically used datasets for pre-training and evaluating SLMs are given in Table~~\ref{table:evaluation_metrics}. These datasets are particularly constructed to satisfy distinct limitations, ensuring robust evaluation across multiple use cases. 

\textbf{Efficient Inference:} For applications where latency and throughput are crucial, models need to create outputs with little delay while retaining high processing rates. This requirement is particularly critical for real-time applications such as question answering, text classification, and natural language understanding. Key datasets in this category include SuperGLUE, SQuAD, TriviaQA, CoQA, and Natural Questions (NQ), which are extensively used for measuring performance in these tasks. 

\textbf{Privacy-Preserving:} With increased concerns about data privacy, privacy-preserving datasets play a critical role in facilitating the development of SLMs that can handle sensitive information securely. PrivacyGLUE ~\cite{havrilla2024noise} applies differential privacy approaches to common tasks such as sentiment analysis, giving a standard for privacy-conscious models. Similarly, anonymized datasets like MIMIC and n2c2\footnote{\url{https://portal.dbmi.hms.harvard.edu/projects/n2c2-nlp/}} offer de-identified clinical notes for medical applications. Federated datasets like LEAF\footnote{\url{https://github.com/TalwalkarLab/leaf}} further enhance privacy by allowing data to remain spread across devices, providing privacy by design using federated learning frameworks. 

\textbf{TinyML and On-device:} Models implemented in resource-constrained situations must run effectively without losing accuracy. Frameworks such as TinyBERT and OpenOrca provide specific datasets and tools for training and evaluation in such circumstances. TinyBERT, a distilled version of BERT, is optimized for both size and speed, making it appropriate for applications with minimal latency needs. Similarly, OpenOrca datasets balance performance with resource limits, permitting the construction of compact and efficient models appropriate for deployment on low-power devices. 

\textbf{Energy-Efficient AI:} In applications where energy consumption is a significant concern, datasets focus on enabling models to optimize energy usage while preserving performance. While specific statistics are not necessarily explicitly connected to this environment, the evaluation generally uses larger datasets to estimate energy efficiency under typical workloads. 

\subsection{Metrics} 

Evaluation measures quantify the performance of SLMs under various limitations, delivering insights into their applicability for particular applications. Table~~\ref{table:evaluation_metrics} organizes these metrics based on their importance to latency, memory, privacy, and energy efficiency. 

\textbf{Latency:} Latency is a critical measure for applications requiring quick feedback, such as chatbots or live translations. Metrics such as inference time assess how quickly a model processes input and creates output, while throughput evaluates the amount of tokens or samples a model can handle within a certain time period. These metrics are particularly significant for large-scale tasks or time-sensitive applications. 

\textbf{Memory:} Memory efficiency is crucial when deploying models in situations with limited memory resources, such as mobile or embedded devices. Metrics such as peak memory usage capture the highest memory consumption during inference, while memory footprint and compression ratio ~\cite{cao2024retaining} evaluate the compactness of a model and the effectiveness of compression approaches. These criteria ensure that models may run efficiently within limited situations without compromising performance. 

\textbf{Privacy:} Protecting sensitive information is critical for models implemented in fields such as healthcare and finance. The privacy budget ~\cite{yu2024privacy} quantifies a model's ability to maintain differential privacy, while noise level ~\cite{havrilla2024understanding} analyzes the trade-off between privacy and accuracy by assessing the amount of noise added to secure data. 

\textbf{Energy Optimization:} As energy efficiency becomes increasingly critical, especially for edge devices and embedded systems, measures like the energy efficiency ratio ~\cite{stojkovic2024dynamollm} provide insights into the energy usage relative to model performance. Additional measures, such as thermal efficiency and idle power consumption ~\cite{patel2024characterizing}, analyze energy use during active processing and idle phases, delivering a full perspective of a model's energy footprint. 

\begin{table*}[ht] 
\centering 
\begin{tabular}{@{}p{4cm}p{3cm}p{5cm}p{4cm}@{}}
\toprule
\textbf{Setting} & \textbf{Constraints} & \textbf{Datasets} & \textbf{Metrics}  \\ 
\midrule 
Efficient Inference & Latency & SuperGLUE, SQuAD, TriviaQA, CoQA, NQ & Inference Time, Throughput \\ 
On-device/Mobile & Memory & TinyBERT, OpenOrca & Peak Memory Usage, Memory Footprint, Compression Ratio \\ 
Privacy-Preserving & Privacy & PrivacyGLUE, MIMIC, n2c2, LEAF & Privacy Budget, Noise Level \\ 
Energy-Efficient AI & Energy Optimization & - & Energy Efficiency Ratio, Thermal Efficiency, Idle Power Consumption \\ 
\bottomrule 
\end{tabular} 
\caption{Overview of settings, constraints, and metrics for evaluating small language models.} 
\label{table:evaluation_metrics} 
\end{table*}


\section{Problems and Future Works}  \label{Solara} 

Despite their growing adoption and versatility, small language models (SLMs) face several major hurdles that impede their full potential. Addressing these problems is crucial for advancing the field and ensuring the reliability and fairness of SLMs across diverse uses. This part discusses some of the most pressing challenges, including hallucination, biases, inference-time energy consumption, and data privacy issues. Each of these open problems offers a chance for future study and development.

\subsection{Problems of Hallucination}

One of the most common problems in both small and large language models is hallucination, which refers to the generation of information that is either truly wrong or irrelevant to the input context. This problem not only undermines user trust but can also have serious effects in high-stakes domains such as healthcare, law, and finance.

Hallucination can be generally grouped into two categories:
\begin{itemize}
    \item \textbf{Factuality Hallucination:} The generated material defies verifiable facts or established knowledge bases.
    \item \textbf{Faithfulness Hallucination:} The result is unrelated to the user's question or the task's requirements, failing to stay contextually relevant.
\end{itemize}

Studies like HallusionBench ~\cite{guan2024hallusionbench} show that bigger models usually exhibit fewer hallucinations, suggesting that increased parameter counts can enhance factual consistency. However, results from the AMBER benchmark suggest that hallucination types and severity change greatly with model size and design. Interestingly, while bigger models may lower the general rate of hallucinations, they are also more likely to produce complicated and nuanced errors, which may be harder to identify and mitigate.

Future study should focus on building benchmarks and assessment measures that account for the variety of hallucination types. Additionally, knowing how model size, training data quality, and design affect hallucination can pave the way for more powerful mitigation methods. 

\subsection{Addressing Bias in Models}

Bias in language models is a longstanding problem that remains even in smaller, more targeted models. Bias often comes from the training data, which may contain social stereotypes, prejudices, or imbalances. This can result in models repeating or even increasing these biases, leading to outcomes that may be unfair or discriminatory.

\textbf{Measuring Bias:} Effective bias reduction starts with solid measurement methods. Several benchmarks have been created for this reason, including:
\begin{itemize}
    \item \textbf{Bias Benchmark for Question Answering (BBQ):} Focuses on finding flaws in question-answering tasks ~\cite{parrish2021bbq}.
    \item \textbf{RealToxicityPrompts:} Measures the tendency of models to create toxic outputs when asked ~\cite{gehman2020realtoxicityprompts}.
    \item \textbf{CrowS-Pairs:} A dataset intended to test for stereotypical links in generated text.
\end{itemize}

\textbf{Impact of Model Size:} Research has shown that larger models tend to display higher levels of bias, as tested across different standards. For instance, the LLaMA series showed greater bias scores on RealToxicityPrompts and StereoSet as model size increased . However, improvements in younger generations, such as LLaMA-2, show that targeted interventions during training can successfully decrease bias levels.

Addressing bias requires a multi-faceted approach, including better management of training data, introduction of fairness-aware loss functions, and model reviews that value equity and inclusion.

\subsection{ Energy-efficient Inference}

Energy economy is a critical issue for applying SLMs in resource-constrained settings, such as mobile devices or embedded systems. These models must balance computing speed with minimal energy consumption to stay effective for real-world uses.

Research using the MELODI standard ~\cite{husom2024price} shows several factors affecting energy efficiency:
\begin{itemize}
    \item GPU-based inference is usually more energy-efficient than CPU-based inference, especially for large-scale jobs.
    \item Devices like computers require significantly more energy during inference compared to specialized hardware.
    \item Response token length is one of the greatest predictors of energy usage, stressing the value of concise outputs.
\end{itemize}

~\cite{stojkovic2024towards} showed that optimization techniques could lower energy usage by up to 20\%, highlighting the possibility for further advancements in this area. Future work should study dynamic adaptation techniques that improve energy use based on application context, device capabilities, and workload characteristics.

\subsection{Data Security and Privacy}

Privacy issues are important when deploying SLMs, especially in applications that handle private user information. The risks connected with data leaks and illegal access must be carefully mitigated to ensure user trust and compliance with privacy regulations.

Privacy problems can be divided into three main areas:
\begin{itemize}
    \item \textbf{Training Data Leakage:} Research by ~\cite{liu2023prompt} showed that data from later steps of pretraining is more sensitive to extraction. The writers suggest that attention mechanisms could be a key factor.
    \item \textbf{System Prompt Leaking:} ~\cite{li2024llm} found risks associated with illegal access to system prompts, which could lead to unintended behavior or misuse of the model.
    \item \textbf{Inference-Time Data Exposure:} Models applied in digital assistants or other apps often access sensitive user data, such as location or health records. Ensuring this data stays safe during inference is important.
\end{itemize}

Mitigation tactics include the use of differential privacy techniques, federated learning frameworks, and strong encryption protocols. However, more study is needed to handle the unique challenges offered by SLMs, especially as they become more integrated into privacy-sensitive applications.

\section{Conclusion} \label{Cryptonix}

The growing importance of Small Language Models (SLMs) lies in their ability to provide efficient and effective solutions across diverse applications, especially in resource-constrained settings. This paper offers a thorough review of SLMs, covering their designs, training methods, and model compression techniques, all of which contribute to their optimization and scalability. By giving an intuitive taxonomy of evaluation measures and summarizing diverse datasets, we have given a method for measuring SLMs' success in different settings. While the breakthroughs in SLM study have been impressive, several basic challenges remain unresolved. Issues such as hallucination, biases, inference-time energy economy, and data protection represent important areas that require further study. We hope that this poll serves as a foundational resource for both academics and practitioners. By finding key areas for innovation and cooperation, we aim to inspire further improvements in the design, training, and implementation of SLMs. As these models continue to develop, they have the potential to revolutionize AI applications by making intelligence systems more accessible, efficient, and trustworthy. Our goal is to see SLMs driving impactful solutions across industries, bridging the gap between technical skills and social needs.

\bibliographystyle{named}
\bibliography{ijcai25}

\end{document}